# Noisy Search with Comparative Feedback

**Shiau Hong Lim and Peter Auer**
Montanuniversität Leoben
Franz-Josef-Straße 18,
8700 Leoben, Austria.

## Abstract

We present theoretical results in terms of lower and upper bounds on the query complexity of noisy search with comparative feedback. In this search model, the noise in the feedback depends on the distance between query points and the search target. Consequently, the error probability in the feedback is not fixed but varies for the queries posed by the search algorithm. Our results show that a target out of $n$ items can be found in $O(\log n)$ queries. We also show the surprising result that for $k$ possible answers per query, the speedup is not $\log k$ (as for $k$-ary search) but only $\log \log k$ in some cases.

## 1 Introduction

We investigate a form of noisy search that arises in information retrieval systems, e.g. content-based image retrieval [Cox et al., 2000]. Consider a system where a user can search for a target item (unknown to the system) by answering a sequence of "queries" generated by the system. As an example, consider a content-based image retrieval system. The system first presents $k$ images to the user. The user then selects one that is the most similar to the target image. Based on the image chosen by the user, the system presents a new set of $k$ images and the user answers by selecting one of these $k$ images. This continues until the target is found by the system. We assume that the user response in each query is probabilistic and depends on the similarities between the presented images and the target image according to a comparative feedback model as proposed in [Cox et al., 2000] and [Auer and Leung, 2009]. The search process is inherently noisy due to the probabilistic nature of the user feedback. We are interested in the performance of such systems in terms of query complexity, i.e the expected number of queries needed before a target is found.

### 1.1 The comparative feedback model

We consider a finite set of $n$ data points $\mathcal{X} = \{x_1, \ldots, x_n\}$ and a search target $T \in \mathcal{X}$. In each query, a set of $k$ ($k \geq 2$) distinct data points $\mathcal{Q} = \{q_1, \ldots, q_k\} \subset \mathcal{X}$ is presented to the user. The search terminates if $T \in \mathcal{Q}$, otherwise the user responds by selecting one of the query points $q_j$, $j \in \{1, \ldots, k\}$.

Let $R \in \{1, \ldots, k\}$ be the random user response (note that we use the indices instead of the query points themselves). The comparative feedback model specifies the probability of choosing a particular response $R = r$ as follows:

$$\Pr(R = r | \mathcal{Q}, T) = \frac{S(T, q_r)}{\sum_{j=1}^{k} S(T, q_j)}$$

where $S(\cdot, \cdot)$ measures the similarity between data points. This implies that query points close to the target are chosen with larger probability than query points far from the target. Given a distance function $d(\cdot, \cdot)$ between data points, we consider two types of similarity measures. The first similarity measure decreases polynomially with increasing distance,

$$S(x, y) = d(x, y)^{-\theta}, \qquad (1)$$

while the second similarity measure decreases exponentially,

$$S(x, y) = \exp\{-\theta d(x, y)\}. \qquad (2)$$

In both cases, $\theta > 0$ is a parameter indicating the user "sharpness". Larger $\theta$ implies that the user favors items closer to the target more intensely.[1]

---

[1]The main difference between the two similarity measures is that for the polynomial similarity measure, the user response depends on the relative differences of the distances to the target, while for the exponential similarity measure the response depends on the absolute differences of the distances. To see this let $k = 2$. Then $\Pr(R = 1 | q_1, q_2, T) =$

## 1.2 Overview of results

The results of our analysis provide both lower bounds and upper bounds for the query complexity. The lower bounds are shown to hold for any algorithm, and the upper bounds provide performance guarantees for actual algorithms that are presented in this paper.

Many formulations of noisy search have been proposed in the literature, each with different assumptions regarding the type of noise/uncertainty present in the user feedback. In the simplest, noise-free case, the search is reduced to standard $k$-ary search, and the query complexity can be as low as $\frac{\log n}{\log k}$ where $n$ is the total number of items and $k$ is the number of possible answers to the queries. In the case of binary search where the user makes a mistake with a fixed probability $p$ in each query, a lower bound of $\frac{\log n + o(\log n)}{1-H(p)}$ [Renyi, 1961] and an upper bound of $\frac{\log n + O(\log \frac{\log n}{\delta})}{1-H(p)}$ [Ben-Or and Hassidim, 2008] are known, where $H(p)$ denotes Shannon's entropy. Similar results for $k$-ary queries with fixed error probability are also available (see [Ben-Or and Hassidim, 2008]). For a survey of a wide range of noisy search problems, see [Pelc, 2002].

The type of uncertainty in the comparative feedback model, however, is unique in the sense that the noise in the user feedback is sensitive to distances among data points. For example, the uncertainty is higher when the query points have relatively similar distances to the target.

In the following section we present our theoretical results, while the proofs are given in a separate section. Most of our results are stated for 1-dimensional data. This might not be realistic for certain data (e.g. images) but highlights the properties of the search problem induced by the comparative feedback model. The main result is that with $k = 2$, $O(\log n)$ query complexity is possible for both the polynomial (Eq. 1) and the exponential (Eq. 2) similarity measures. This result carries over also to data of arbitrary dimension, but with a possible cost that is exponential in the dimension. Surprisingly, we can show that larger $k$ helps more for the exponential similarity measure (by a factor $\log k$ as expected from $k$-ary search), while the improvement for the polynomial similarity measure (Eq. 1) is shown to be at most by a factor $\log \log k$ in some cases.

## 2 Results

We assume that the user model and its parameters are known to the search algorithm.

Except for the results in Section 2.2 we also assume that $\mathcal{X} \subset \mathbb{R}$ and that the distance between data points is measured as $d(x_i, x_j) = |x_i - x_j|$.

Since the search algorithm knows the user model, it can maintain the posterior target distribution in respect to the queries and user responses so far. We denote the posterior probability of data point $x_i$ being the target by $a_i$ and we use $\boldsymbol{a}$ to denote the vector $(a_1, \ldots, a_n)$. After receiving a user response $R = r$ to a query $\mathcal{Q}$, this posterior is updated as

$$\begin{aligned} a'_i \leftarrow b_{i,r} &:= \Pr(T = x_i | \mathcal{Q}, R = r, \boldsymbol{a}) \\ &= \frac{\Pr(T = x_i | \boldsymbol{a}) \Pr(R = r | \mathcal{Q}, T = x_i)}{\Pr(R = r | \mathcal{Q}, \boldsymbol{a})} \\ &= \frac{a_i p_{i,r}}{A_r} \end{aligned}$$

where

$$p_{i,r} = \Pr(R = r | \mathcal{Q}, T = x_i) = \frac{S(x_i, q_r)}{\sum_{j=1}^{k} S(x_i, q_j)}$$

and

$$A_r = \Pr(R = r | \mathcal{Q}, \boldsymbol{a}) = \sum_{i=1}^{n} a_i p_{i,r}.$$

When $\mathcal{X} \subset \mathbb{R}$, we assume $x_1 < \cdots < x_n$ and denote by $c_i$ the cumulative probabilities

$$c_i = \sum_{i'=1}^{i} a_{i'}.$$

The quantiles of the cumulative probability are denoted by $I(p)$ where $I(p)$ is the index such that

$$c_{I(p)-1} < p \leq c_{I(p)}.$$

We will use $\boldsymbol{A}$ for $(A_1, \ldots, A_k)$ and $\boldsymbol{p}_i$ for $(p_{i,1}, \ldots, p_{i,k})$. Finally, we denote the entropy function by

$$H(\boldsymbol{a}) = -\sum_{i=1}^{n} a_i \log a_i \quad ,$$

and the KL-divergence by

$$D(\boldsymbol{a} || \boldsymbol{a}') = \sum_{i=1}^{n} a_i \log \frac{a_i}{a'_i} \quad .$$

We will use base-2 logarithms throughout. Also, for scalar $p$ and $p'$, we define $H(p) = H(p, 1-p)$, and similarly $D(p||p') = D((p, 1-p)||(p', 1-p'))$.

---

$1/(1+[\frac{d(q_1,T)}{d(q_2,T)}]^\theta)$ for the polynomial similarity measure and $\Pr(R = 1 | q_1, q_2, T) = 1/(1+\exp\{\theta[d(q_1,T)-d(q_2,T)]\})$ for the exponential similarity measure.

## 2.1 Efficient search for the polynomial similarity measure

In this section we show, for $\mathcal{X} \subset \mathbb{R}$, how query complexity $O(\log n)$ can be achieved in the comparative feedback model with polynomial similarity measure for $k = 2$. The initial uncertainty about the target is $\log n$ when measured by entropy. To guarantee at most $O(\log n)$ query iterations, we need to select query points such that a constant information gain is guaranteed. This can be achieved in the following way:

- Calculate the quantiles $i_s = I(s/4)$ for $s = 0, \ldots, 4$ ($i_0 = 1$, $i_4 = n$), and consider the resulting 4 intervals (see Fig. 1 for an example, where $d_1, \ldots, d_4$ mark the intervals).

- From these four intervals find the one with the smallest length.

- Query the endpoints of that interval adjacent to the smallest interval, which does not contain $x_1$ or $x_n$.

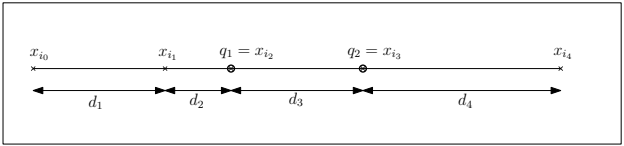

Figure 1: Selection of query points; $d_2$ is the smallest interval, $d_3$ is adjacent to $d_2$ and does not contain $x_1$ or $x_n$

Figure 1 shows an example for the selection of the query points. The rationale for this algorithm can be seen by this example: for any data point $x$ in interval $d_2$ we have $|x - q_1|/|x - q_2| \leq 1/2$, while for any data point $x$ in interval $d_4$ we have $|x - q_1|/|x - q_2| \geq 1$. Thus a user response $R = 1$ significantly indicates that the target is rather in $d_2$ than in $d_4$, while a user response $R = 2$ mildly indicates that the target is rather in $d_4$ than in $d_2$. This is sufficient to prove the following theorem, the proof is given in Section 3.2.

**Theorem 1.** *For the comparative feedback model with polynomial similarity measure, the expected number of queries of the above algorithm is at most*

$$\frac{4 \log_2 n}{G_\rho} + 4,$$

where

$$G_\rho = \tfrac{1}{4}\left(D(\rho\|\phi) + D(\tfrac{1}{2}\|\phi)\right) > 0$$

*with $\rho = \frac{2^\theta}{1+2^\theta}$ and $\phi = \frac{1}{4} + \frac{1}{2}\rho$. The expectation is taken over targets selected uniformly at random and the random responses of the user.*

## 2.2 Search in a $D$-dimensional space

The next theorem generalizes Theorem 1 to $D$-dimensional data points. The drawback is that the constant in the upper bound depends exponentially on the dimension. It can indeed be shown that this is necessary if distances between data points are measured by the $\infty$-norm.

We assume that distances are measured by some $p$-norm. To mimic the construction of Theorem 1, we need a strategy for choosing query points such that there are two "well-separated" groups of data points with significant probability mass. This can be achieved as follows:

- Let $B$ be the smallest $\|\cdot\|_p$-ball with probability mass $\sum_{i:x_i \in B} a_i \geq c$ for some positive constant $c$ (to be determined later). In case of ties, the $\|\cdot\|_p$-ball with the largest total probability mass is chosen.

- Choose an arbitrary point in $B$ as the first query point $q_1$.

- Let $\lambda$ be the radius of $B$. Let $B'$ be the $\|\cdot\|_p$-ball centered on $B$ with radius $7\lambda$.

- Choose the second query point $q_2$ as the data point not in $B'$ that is closest to $q_1$.

By the above construction we have for all $x \in B$, $\|x - q_1\|_p \leq 2\lambda$ and $\|x - q_2\|_p \geq \|q_2\|_p - \|x\|_p \geq 6\lambda$, such that

$$\|x - q_1\|_p \leq \|x - q_2\|_p/3,$$

and for all $x \notin B'$ we have $\|x - q_2\|_p \leq \|x - q_1\|_p + \|q_1 - q_2\|_p \leq 2\|x - q_1\|_p$ such that

$$\|x - q_1\|_p \geq \|x - q_2\|_p/2.$$

Since $(14D)^D$ $\|\cdot\|_p$-balls of radius $\lambda/2$ are sufficient to cover $B'$ ([2]), the probability mass of $B'$ is at most $(14D)^D c$ (otherwise there would be a $\|\cdot\|_p$-ball of radius $\lambda/2$ with probability mass at least $c$). Thus for $c = (14D)^{-D}/2$ the probability mass of the data points not in $B'$ is at least $1/2$. This is sufficient to prove, similarly to Theorem 1, the following theorem. The detailed proof is omitted.

**Theorem 2.** *For the comparative feedback model with polynomial similarity measure and $D$-dimensional data, the expected number of queries of the above algorithm is at most $O\left((14D)^D \log n\right)$.*

---

[2]This can be seen by covering $B'$ with $\|\cdot\|_\infty$-balls of radius $\frac{\lambda}{2D}$, and then covering each of these $\|\cdot\|_\infty$-balls by a $\|\cdot\|_p$-ball of radius $\lambda/2$.

### 2.3 For the polynomial similarity measure large $k$ may not help much

Using $k$ query points, one could expect that the number of query iterations — similarly as for $k$-ary search — can be reduced by a factor of $\log k$. Surprisingly, we can show that for some search problems this is not the case in the comparative feedback model with polynomial similarity measure.

**Theorem 3.** *For the comparative feedback model with polynomial similarity measure, there are data points $x_1 < \cdots < x_n \in \mathbb{R}$ such that the expected number of queries for any search algorithm is at least*

$$\frac{\log n - \log(2k)}{2 \log \log k + 4} = \Omega\left(\frac{\log n}{\log \log k}\right)$$

*when the target is selected uniformly at random from $\{x_1, \ldots, x_n\}$.*

We will choose increasingly distant data points, $x_{i+1} - x_i \gg x_i - x_{i-1}$, such that the similarity of $x_i$ to data points with smaller index is not much smaller than the similarity to data points with larger index, $S(x_i, x_1) \geq S(x_i, x_{i+1})/2$. The idea behind this is that then for any query $q_1 < \cdots < q_k$ and any target $T = x_i$, the probability of the user response can be bounded by $\Pr\{R = j | T = x_i\} \leq 2/j$. Thus query responses with large $j$ are rather unlikely for any target point, which ensures that the expected information gain per query response is only $\log \log k$, in contrast to information gain $\log k$ for $k$-ary search with deterministic responses. The full calculation leading to Theorem 3 is given in Section 3.3.

### 2.4 The exponential similarity measure allows $k$-ary search

In contrast to the previous section, we show in this section that for the exponential similarity measure it is always possible to find the target within $O\left(\frac{\log n}{\log k}\right)$ queries. This is achieved by the following selection algorithm: the algorithm subsequently chooses $k$ disjoint intervals $\mathcal{I}_1, \ldots, \mathcal{I}_k$ of minimal length such that[3]

- each interval has probability mass at least

$$c = \frac{1}{14k - 12},$$

i.e. $\sum_{i: x_i \in \mathcal{I}_j} a_i \geq c$,

- and the distance between two intervals $\mathcal{I}_j = [x_j, y_j]$ and $\mathcal{I}_{j'} = [x_{j'}, y_{j'}]$ is at least the length

---

[3]In case the construction fails, it is easy to show that there must already exist data points with total probability mass at least $1/2$. This will be covered in the proof.

of the smaller interval, i.e. $x_{j'} - y_j \geq \min\{y_j - x_j, y_{j'} - x_{j'}\}$ for $y_j < x_{j'}$.

The algorithm selects one endpoint from each interval as a query point. For each interval $\mathcal{I}_j = [x_j, y_j]$, if the first half of the interval $[x_j, \frac{x_j + y_j}{2}]$ contains more probability mass than the second half $[\frac{x_j + y_j}{2}, y_j]$ then $x_j$ is selected as the query point, otherwise $y_j$ is selected.

Let denote by $\mathcal{X}'$ all data points in these intervals that are closer to the query point of the respective interval than to any other query point. By construction, the probability mass of these points is at least $c/2$ in each interval. For each $x_i \in \mathcal{X}'$ let $r_i$ be the index of the corresponding query point (i.e. the query point is $q_{r_i}$). Furthermore, let $\delta_0$ be the minimal distance between any two data points. Then it can be shown that there is a constant $\beta > 0$ depending on $\delta_0$ and the user sharpness $\theta$ (but not on $k$), such that for any $x_i \in \mathcal{X}'$,

$$p_{i, r_i} \geq \beta.$$

Together with the observation that the probability mass of $\mathcal{X}'$ is at least $ck/2 \geq 1/28$, this is sufficient to show an information gain of $\Omega(\log k)$ in each query. This gives the following result.

**Theorem 4.** *For the comparative feedback model with exponential similarity measure and data points $x_i \in \mathbb{R}$, the expected number of queries for the above selection algorithm is at most $O\left(\frac{\log n}{\log k}\right)$.*

The proofs are provided in Section 3.4.

## 3 Proofs

### 3.1 Preliminaries

**Lemma 1.** *The expected information gain in each query is given by:*

$$\mathrm{E}\big[H(\boldsymbol{a}) - H(\boldsymbol{a}')\big] = \sum_{i=1}^{n} a_i D(\boldsymbol{p}_i \| \boldsymbol{A})$$

*Proof.* After each query, the expected entropy of the posterior is given by:

$$\begin{aligned}
\mathrm{E}\big[H(\boldsymbol{a}')\big] &= \sum_{j=1}^{k} \Pr(R = j) H(b_{1,j}, \ldots, b_{n,j}) \\
&= -\sum_{j=1}^{k} A_j \sum_{i=1}^{n} b_{i,j} \log b_{i,j} \\
&= -\sum_{j=1}^{k} A_j \sum_{i=1}^{n} \frac{a_i p_{i,j}}{A_j} \log \frac{a_i p_{i,j}}{A_j}
\end{aligned}$$

$$= -\sum_{j=1}^{k}\sum_{i=1}^{n} a_i p_{i,j}\left(\log a_i + \log \frac{p_{i,j}}{A_j}\right)$$

$$= H(\boldsymbol{a}) - \sum_{i=1}^{n} a_i D(\boldsymbol{p}_i || \boldsymbol{A})$$

Rearranging the terms completes the proof. □

**Lemma 2.** *Let $\boldsymbol{\phi}_1, \ldots, \boldsymbol{\phi}_l$ be any set of $l$ probability distributions on the $k$ query points and $\alpha_1, \ldots, \alpha_l$ be any set of $l$ positive weights. Then, for any distribution $\boldsymbol{r}$ on the $k$ query points,*

$$\sum_{i=1}^{l} \alpha_i D(\boldsymbol{\phi}_i || \boldsymbol{r}) \geq \sum_{i=1}^{l} \alpha_i D(\boldsymbol{\phi}_i || \bar{\boldsymbol{\phi}})$$

*where*

$$\bar{\boldsymbol{\phi}} = \frac{\sum_{i=1}^{l} \alpha_i \boldsymbol{\phi}_i}{\sum_{i=1}^{l} \alpha_i} \quad .$$

*Proof.*

$$\sum_{i=1}^{l} \alpha_i D(\boldsymbol{\phi}_i || \boldsymbol{r})$$

$$= \sum_{i=1}^{l} \alpha_i \sum_{j=1}^{k} \phi_{i,j} \log \left(\frac{\bar{\phi}_j}{r_j}\right)\left(\frac{\phi_{i,j}}{\bar{\phi}_j}\right)$$

$$= \sum_{j=1}^{k} \left(\sum_{i=1}^{l} \alpha_i \phi_{i,j}\right) \log \frac{\bar{\phi}_j}{r_j} + \sum_{i=1}^{l} \alpha_i \sum_{j=1}^{k} \phi_{i,j} \log \frac{\phi_{i,j}}{\bar{\phi}_j}$$

$$= \left(\sum_{i=1}^{l} \alpha_i\right) \sum_{j=1}^{k} \bar{\phi}_j \log \frac{\bar{\phi}_j}{r_j} + \sum_{i=1}^{l} \alpha_i \sum_{j=1}^{k} \phi_{i,j} \log \frac{\phi_{i,j}}{\bar{\phi}_j}$$

$$= \left(\sum_{i=l}^{l} \alpha_i\right) D(\bar{\boldsymbol{\phi}} || \boldsymbol{r}) + \sum_{i=1}^{l} \alpha_i D(\boldsymbol{\phi}_i || \bar{\boldsymbol{\phi}})$$

$$\geq \sum_{i=1}^{l} \alpha_i D(\boldsymbol{\phi}_i || \bar{\boldsymbol{\phi}})$$

where the last inequality is due to non-negativity of relative entropy. □

**Lemma 3.** *Let $\Phi \subset \{1, \ldots, n\}$ be any subset of indices for the data points. The expected information gain in any query is at least*

$$\sum_{i \in \Phi} a_i D(\boldsymbol{p}_i || \boldsymbol{\phi})$$

*where $\boldsymbol{\phi} = \frac{\sum_{i \in \Phi} a_i \boldsymbol{p}_i}{\sum_{i \in \Phi} a_i}$.*

*Proof.* Apply Lemma 2 to Lemma 1. □

### 3.2 Proof of Theorem 1

**Lemma 4.** *Let $\Phi, \Psi \subset \{1, \ldots, n\}$ be two disjoint sets of indices such that $\sum_{i \in \Phi} a_i \geq \alpha$, $\sum_{i \in \Psi} a_i \geq \beta$, and*

$$(\min_{i \in \Phi} p_i) = p > q = (\max_{i \in \Psi} p_i) \quad .$$

*Let*

$$\bar{p} = \frac{\sum_{i \in \Phi \cup \Psi} a_i p_i}{\sum_{i \in \Phi \cup \Psi} a_i} \quad \text{and} \quad \mu = \frac{\alpha p + \beta q}{\alpha + \beta} \quad .$$

*Then*

$$\sum_{i \in \Phi \cup \Psi} a_i D(p_i || \bar{p}) \geq \alpha D(p||\mu) + \beta D(q||\mu) \quad .$$

*Proof.* Let $a_\Phi = \sum_{i \in \Phi} a_i$, $a_\Psi = \sum_{i \in \Psi} a_i$,

$$p_\Phi = \sum_{i \in \Phi} \frac{a_i}{a_\Phi} p_i \quad \text{and} \quad p_\Psi = \sum_{i \in \Psi} \frac{a_i}{a_\Psi} p_i \quad .$$

Note that

$$\bar{p} = \frac{a_\Phi p_\Phi + a_\Psi p_\Psi}{a_\Phi + a_\Psi} \quad .$$

We have

$$\sum_{i \in \Phi \cup \Psi} a_i D(p_i || \bar{p})$$

$$= \sum_{i \in \Phi} a_i D(p_i || \bar{p}) + \sum_{i \in \Psi} a_i D(p_i || \bar{p})$$

$$= a_\Phi \sum_{i \in \Phi} \frac{a_i}{a_\Phi} D(p_i || \bar{p}) + a_\Psi \sum_{i \in \Psi} \frac{a_i}{a_\Psi} D(p_i || \bar{p})$$

$$\geq a_\Phi D(p_\Phi || \bar{p}) + a_\Psi D(p_\Psi || \bar{p})$$

where the inequality is due to convexity of $D$.

Let $f(a_\Phi, a_\Psi) = a_\Phi D(p_\Phi || \bar{p}) + a_\Psi D(p_\Psi || \bar{p})$ be a function of $a_\Phi$ and $a_\Psi$. It can be shown, by taking derivatives, that $f(a_\Phi, a_\Psi) \geq f(\alpha, \beta)$.

Assume now that $a_\Phi = \alpha$ and $a_\Psi = \beta$. Let $g(p_\Phi, p_\Psi) = \alpha D(p_\Phi || \bar{p}) + \beta D(p_\Psi || \bar{p})$ be a function of $p_\Phi$ and $p_\Psi$. Taking derivatives with respect to $p_\Phi$ and $p_\Psi$, and using the fact that $p > q$, it is straightforward to show that the minimum is at $g(p, q)$. □

**Lemma 5.** *In each query iteration the algorithm of Section 2.1 achieves at least*

$$G_\rho = \tfrac{1}{4}\left(D(\rho||\phi) + D(\tfrac{1}{2}||\phi)\right)$$

*expected information gain where $\rho = \frac{2^\theta}{1+2^\theta}$ and $\phi = \tfrac{1}{4} + \tfrac{1}{2}\rho$, if all four intervals considered by the algorithm have non-zero length. Furthermore, $G_\rho > 0$.*

*Proof.* Lemma 3 allows us to ignore the contribution of subsets of data points in analyzing the information gain. In particular, we will focus attention onto only two of the four intervals of data points, the smallest interval and another interval such that the two query points are between these two intervals. Without loss of generality we assume the configuration of Figure 1, such that these intervals are $d_2$ and $d_4$.

The key property of all data points in interval $d_2$ is that the distance to $q_1$ is at most half the distance to $q_2$. Therefore, for any $x_i \in d_2$,

$$p_{i,1} = \frac{S(x_i, q_1)}{S(x_i, q_1) + S(x_i, q_2)} = \frac{|x_i - q_2|^\theta}{|x_i - q_1|^\theta + |x_i - q_2|^\theta}$$
$$\geq \frac{2^\theta}{1 + 2^\theta} = \rho > \frac{1}{2}.$$

On the other hand, for any $x_i \in d_4$, $p_{i,1} < \frac{1}{2}$. Since $\sum_{x_i \in d_2} a_i \geq 1/4$ and $\sum_{x_i \in d_4} a_i \geq 1/4$, applying Lemma 3 and Lemma 4 (setting $\alpha = \beta = 1/4$, $p = \rho$ and $q = 1/2$), the expected information gain is at least

$$\sum_{i \in I_{\sigma_1} \cup I_{\sigma_2}} a_i D(p_i \| \bar{p}) \geq \frac{1}{4} D(\rho \| \phi) + \frac{1}{4} D(\frac{1}{2} \| \phi) .$$

Finally, since $\theta > 0$, $\frac{1}{2} < \phi < \rho$, and hence $D(\rho \| \phi) > 0$ and $D(\frac{1}{2} \| \phi) > 0$. □

*Proof of Theorem 1.* From Lemma 5 we get that the expected number of query iterations until one of the considered intervals is of zero length, is at most $(\log n)/G_\rho$. If there is a zero length interval, then there is a data point $x_i$ with $a_i \geq 1/4$ that is selected as one of the query points $q_1$ or $q_2$. Thus in this case the search terminates with probability at least $1/4$.

Now let $\tau$ be an upper bound on the expected total number of query iterations. Then the above reasoning gives the following recursion,

$$\tau \leq \frac{1}{4}\left[\frac{\log n}{G_\rho} + 1\right] + \frac{3}{4}\left[\frac{\log n}{G_\rho} + 1 + \tau\right],$$

since the search essentially might restart when the query point from a zero length interval is not the target. Solving for $\tau$ gives

$$\tau \leq \frac{4 \log n}{G_\rho} + 4.$$

□

### 3.3 Proof of Theorem 3

The following is a suitable choice of the data points: for all $i > 1$,

$$x_{i+1} = \frac{x_i - x_1 \cdot 2^{-1/\theta}}{1 - 2^{-1/\theta}}$$

such that

$$\left(\frac{x_{i+1} - x_i}{x_{i+1} - x_1}\right)^\theta = \frac{1}{2},$$

where $\theta$ is the parameter in the polynomial similarity measure, $S(x, y) = |x - y|^{-\theta}$.

**Lemma 6.** *With the above choice of data points,*

$$S(x, x_i) \leq 2S(x, x_1)$$

*for any data points $x \neq x_i$.*

*Proof.* By construction we have $S(x_{i+1}, x_i) = 2S(x_{i+1}, x_1)$. Since $\frac{S(x,x_i)}{S(x,x_1)}$ is decreasing in $x$, the statement holds for any $x \geq x_{i+1}$. Furthermore, for $x \leq x_{i-1}$, $S(x, x_i) = S(x_i, x) \leq 2S(x_i, x_1) \leq 2S(x, x_1)$. □

**Lemma 7.** *With the above choice of data points, for any query $q_1 < \cdots < q_k$ and any data point $x \notin \{q_1, \ldots, q_k\}$, the probability of choosing the $j$-th query point is bounded by*

$$\Pr(R = j | T = x) \leq \frac{2}{j} .$$

*Proof.* Let $\ell$ be the index of the query points such that $q_\ell < x < q_{\ell+1}$ and assume that $\ell \leq j$. Then

$$\Pr(R = j | T = x) = \frac{S(x, q_j)}{\sum_{j'=1}^{k} S(x, q_{j'})}$$
$$= \frac{S(x, q_j)}{\sum_{j'=1}^{\ell} S(x, q_{j'}) + \sum_{j'=\ell+1}^{k} S(x, q_{j'})}$$
$$\leq \frac{S(x, q_j)}{\sum_{j'=1}^{\ell} S(x, q_{j'}) + \sum_{j'=\ell+1}^{j} S(x, q_{j'})}$$
$$\leq \frac{S(x, q_j)}{\ell \cdot S(x, x_1) + (j - \ell) \cdot S(x, q_j)}$$
$$\leq \frac{S(x, q_j)}{j \cdot \min\{S(x, x_1), S(x, q_j)\}}$$
$$\leq \frac{2}{j}$$

by Lemma 6. If $\ell > j$ then

$$\frac{S(x, q_j)}{\sum_{j'=1}^{k} S(x, q_{j'})} \leq \frac{S(x, q_j)}{\sum_{j'=1}^{j} S(x, q_{j'})} \leq \frac{S(x, q_j)}{j \cdot S(x, x_1)} \leq \frac{2}{j}$$

again by Lemma 6. □

Let $R_t \in \{1, \ldots, k\}$ be the random user response to the $t$-th query and let $\omega$ denote a possible randomization of the search algorithm. Then, given a particular sequence of user responses $R_1 = j_1, R_2 = j_2, \ldots, R_t = j_t$, the set of query points chosen by the algorithm for the $(t+1)$-th query is given by a function $Q(j_1, j_2, \ldots, j_t | \omega)$. Let $Q_t$ be the random variable $Q_t = Q(R_1, \ldots, R_{t-1} | \omega)$. We assume that the target $T \notin Q_t$ if $T \in Q_\tau$ for some $\tau < t$.

**Lemma 8.** *If the target $T$ is chosen uniformly at random from the choice of data points above, then for $k \geq 3$,*

$$\Pr(T \in Q_t) \leq \frac{k}{n}(4 \log k)^{t-1}.$$

*Proof.*

$$\begin{aligned}
&\Pr(T \in Q_t) \\
&= \frac{1}{n} \sum_\omega \sum_x \Pr(T \in Q_t | T = x, \omega) \Pr(\omega) \\
&= \frac{1}{n} \sum_\omega \sum_x \sum_{j_1,\ldots,j_{t-1}} \Pr(R_1 = j_1, \ldots \\
&\quad , R_{t-1} = j_{t-1}, T \in Q_t | T = x, \omega) \Pr(\omega) \\
&= \frac{1}{n} \sum_\omega \sum_x \sum_{j_1,\ldots,j_{t-1}} \Pr(R_1 = j_1, \ldots \\
&\quad , R_{t-1} = j_{t-1}, x \in Q(j_1, \ldots, j_{t-1} | \omega) | T = x, \omega) \Pr(\omega) \\
&= \frac{1}{n} \sum_\omega \sum_{j_1,\ldots,j_{t-1}} \sum_{x \in Q(j_1,\ldots,j_{t-1}|\omega)} \\
&\quad \Pr(R_1 = j_1, \ldots, R_{t-1} = j_{t-1} | T = x, \omega) \Pr(\omega) \\
&\leq \frac{1}{n} \sum_\omega \sum_{j_1,\ldots,j_{t-1}} \sum_{x \in Q(j_1,\ldots,j_{t-1}|\omega)} \\
&\quad \left(\frac{2}{j_1}\right) \cdots \left(\frac{2}{j_{t-1}}\right) \Pr(\omega) \qquad (3) \\
&= \frac{k}{n} \sum_{j_1,\ldots,j_{t-1}} \left(\frac{2}{j_1}\right) \cdots \left(\frac{2}{j_{t-1}}\right) \\
&\leq \frac{k}{n}(4 \log k)^{t-1}
\end{aligned}$$

for $k \geq 3$. Inequality (3) follows from Lemma 7 and the fact that the user responses are independent given the target. □

*Proof of Theorem 3.* Using Lemma 8, we have

$$\sum_{t=1}^\tau \Pr(T \in Q_t) \leq \frac{k}{n} \frac{(4 \log k)^\tau - 1}{4 \log k - 1} \leq \frac{k}{n}(4 \log k)^\tau \leq \frac{1}{2}$$

for

$$\tau \leq \frac{\log \frac{n}{2k}}{\log \log k + 2},$$

and the theorem follows from Markov's inequality. □

### 3.4 Proof of Theorem 4

The query construction of the selection algorithm fails only if there are $k$ data points with total probability mass $\geq 1/2$. We will deal with this case later. For now we assume that the construction succeeds.

**Lemma 9.** *For any $x_i \in \mathcal{X}'$,*

$$p_{i,r_i} \geq \beta$$

*where*

$$\beta = \frac{1}{1 + 2e^{-\theta \delta_0}(1 + \frac{1}{\theta \delta_0})}.$$

*Proof.* For $x_i \in \mathcal{X}'$,

$$\begin{aligned}
p_{i,r_i} &= \Pr(R = r_i | T = x_i) \\
&= \frac{S(x_i, q_{r_i})}{\sum_{j=1}^k S(x_i, q_j)} \\
&= \frac{1}{1 + \sum_{j \neq r_i} e^{-\theta(|x_i - q_j| - |x_i - q_{r_i}|)}} \\
&\geq \frac{1}{1 + \sum_{j \neq r_i} e^{-\theta(|j - r_i|\delta_0)}} \qquad (4) \\
&\geq \frac{1}{1 + 2\sum_{j'=1}^{(k-1)/2} e^{-j'\theta\delta_0}} \qquad (5) \\
&\geq \frac{1}{1 + 2e^{-\theta\delta_0}(1 + \frac{1}{\theta\delta_0})} \qquad (6)
\end{aligned}$$

For inequality (4), by construction, the distance between any two adjacent query points is at least $\delta_0$ and the distance between any two query points $j'$ intervals apart is $j'\delta_0$. In inequality (5), we rewrite the indices and use the worst case where $r_i$ is the query point in the center-most interval, with $(k-1)/2$ query points on each side of $q_{r_i}$. The inequality (6) is obtained using the integral bound. □

**Lemma 10.** *For any query constructed as described in Section 2.4, the expected information gain is $\Omega(\log k)$.*

*Proof.* Using Lemma 3 we find that the expected information gain is at least

$$\sum_{i:x_i \in \mathcal{X}'} a_i D(\boldsymbol{p}_i \| \boldsymbol{\phi}) \qquad (7)$$

where

$$\boldsymbol{\phi} = \frac{\sum_{i:x_i \in \mathcal{X}'} a_i \boldsymbol{p}_i}{\sum_{i:x_i \in \mathcal{X}'} a_i}.$$

Given the constraint of Lemma 9, (7) is minimized for $p_{i,r_i} = \beta$, $p_{i,r} = (1-\beta)/(k-1)$ for $r \neq r_i$, and $\phi_r = 1/k$.[4] Thus

$$\begin{aligned}
D(\boldsymbol{p}_i \| \boldsymbol{\phi}) &= \beta \log(\beta k) + (1-\beta) \log \frac{(1-\beta)k}{k-1} \\
&= \beta \log k + (1-\beta) \log\left(1 + \frac{1}{k-1}\right) - H(\beta) \\
&= \beta \log k + O(1).
\end{aligned}$$

Since $\sum_{i:x_i \in \mathcal{X}'} a_i \geq 1/28$, this completes the proof. □

---
[4]The minimum can be found using the method of Lagrange multipliers.

*Proof of Theorem 4.* Using Lemma 10, the proof proceeds similarly to the proof of Theorem 1, observing that if the construction of the selection algorithm fails, then the search terminates with probability $\geq 1/2$. □

## 4 Empirical testing of the feedback model

A question regarding the comparative feedback model is whether it adequately models the uncertainty of user response in real-world applications. We performed statistical goodness-of-fit tests on actual user feedback in image search tasks where distances are measured by Euclidean distance in the image feature space [Auer et al., 2011]. In most of the tasks, the model with polynomial similarity measure fits the data very well while in contrast, the model with exponential similarity measure can be rejected with high confidence. We believe that the distance metric plays a crucial role here and more conclusive results require further investigations.

The results in this paper assume that the model parameter $\theta$ is known to the algorithm, which is unlikely in practice. However, empirical evidence suggests that search performance degrades rather gracefully under parameter mismatch. To illustrate this, Figure 2 shows the query performance with respect to varying $\theta$ assumed by the algorithm while the true $\theta$ is indicated by a cross.

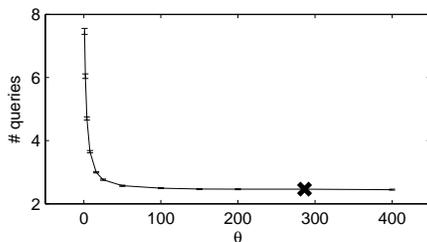

Figure 2: Performance under parameter mismatch

## 5 Open problems

There are some open questions concerning the necessary number of queries, that we would like to address in future work:

- Can the exponential dependency on the dimension be avoided for $D$-dimensional data? Initial progress seems to indicate that for distances measured by $p$-norms with $p < \infty$, the dependency on the dimension is only polynomial, while for the $\infty$-norm the dependency is exponential.

- Is there an algorithm that — for 1-dimensional data — guarantees $O\left(\frac{\log n}{\log \log k}\right)$ queries for any set of data points? We already can show that for data points as chosen in Section 2.3, indeed only $O\left(\frac{\log n}{\log \log k}\right)$ queries are necessary.

- Can our analysis — which assumes a uniform (or known) prior distribution on the possible targets and known parameters of the feedback model — be extended to an unknown prior distribution and unknown parameters of the feedback model?

## Acknowledgements

The research leading to these results has received funding from the European Community's Seventh Framework Programme (FP7/2007-2013) under grant agreement n° 231495 (CompLACS), n° 216886 (PASCAL2), n° 216529, Personal Information Navigator Adapting Through Viewing, PinView, and the Austrian Federal Ministry of Science and Research.

## References

[Auer et al., 2011] Auer, P., Glowacka, D., Leung, A., Lim, S. H., Medlar, A., and Shawe-Taylor, J. (2011). Exploration-exploitation trade-offs with delayed feedback. PinView Deliverable D4.3, www.pinview.eu/deliverables.

[Auer and Leung, 2009] Auer, P. and Leung, A. (2009). Relevance feedback models for content-based image retrieval. *Multimedia Analysis, Processing and Communications*.

[Ben-Or and Hassidim, 2008] Ben-Or, M. and Hassidim, A. (2008). The bayesian learner is optimal for noisy binary search (and pretty good for quantum as well). In *Proceedings of the 2008 49th Annual IEEE Symposium on Foundations of Computer Science*, pages 221–230, Washington, DC, USA. IEEE Computer Society.

[Cox et al., 2000] Cox, I., Miller, M., Minka, T., Papathomas, T., and Yianilos, P. (2000). The Bayesian image retrieval system, PicHunter: theory, implementation, and psychophysical experiments. *IEEE Trans. Image Processing*, 9(1):20–37.

[Pelc, 2002] Pelc, A. (2002). Searching games with errors—fifty years of coping with liars. *Theor. Comput. Sci.*, 270:71–109.

[Renyi, 1961] Renyi, A. (1961). On a problem of information theory. *MTA Mat. Kut. Int. Kozl.*, 6B:505–516.